\documentclass[letterpaper, 10 pt, conference]{ieeeconf}  
\usepackage{comment}
\usepackage{graphicx}
\usepackage{subcaption}

\IEEEoverridecommandlockouts                              

\title{\LARGE \bf
Kitchen Artist: Precise Control of Liquid Dispensing \\for Gourmet Plating

}

\author{Hung-Jui Huang$^{1}$, Jingyi Xiang$^{2}$, and Wenzhen Yuan$^{2}$
\thanks{$^{1}$Hung-Jui Huang is with Carnegie Mellon University, Pittsburgh, PA, USA 
        {\tt\small hungjuih@andrew.cmu.edu}}%
\thanks{$^{2}$Jingyi Xiang and Wenzhen Yuan are with University of Illinois Urbana-Champaign, Champaign, IL, USA
        {\tt\small \{jingyix4, yuanwz\}@illinois.edu }}%
}

\begin{document}

\maketitle
\thispagestyle{empty}
\pagestyle{empty}

\begin{abstract}
Manipulating liquid is widely required for many tasks, especially in cooking. A common way to address this is extruding viscous liquid from a squeeze bottle. In this work, our goal is to create a sauce plating robot, which requires precise control of the thickness of squeezed liquids on a surface. Different liquids demand different manipulation policies.  We command the robot to tilt the container and monitor the liquid response using a force sensor to identify liquid properties. Based on the liquid properties, we predict the liquid behavior with fixed squeezing motions in a data-driven way and calculate the required drawing speed for the desired stroke size. This open-loop system works effectively even without sensor feedback. Our experiments demonstrate accurate stroke size control across different liquids and fill levels. We show that understanding liquid properties can facilitate effective liquid manipulation. More importantly, our dish garnishing robot has a wide range of applications and holds significant commercialization potential.

\end{abstract}

\section{Introduction} \label{introduction}

Liquids are common substances in many environments, with kitchens being a prime example. Human manipulation of liquids, including dispensing them into different containers in specific patterns, is a common practice. However, effectively handling liquids poses significant challenges. Liquids are elusive to perceive and control, and even slight adjustments in holding force or position can have a substantial impact on their behavior. Furthermore, the diverse and often imperceptible properties of liquids can greatly affect their motion patterns. Consequently, it comes as no surprise that liquid handling remains a formidable challenge for robots. While some research has explored liquid perception, such as liquid classification \cite{Guler2014} or property recognition \cite{Matl2019} \cite{Huang2022}, the realm of liquid manipulation has seen limited progress. Robots have only mastered pouring \cite{Schenck2017} \cite{Liang2019} \cite{Yamaguchi2016}, a task that primarily focuses on low-viscosity liquids where liquid properties have a negligible influence.

In this work, we aim to enable a robot to garnish dishes by drawing with different sauces. This is done by extruding the sauces from a bottle in a well-controlled way. A central challenge in this task involves executing smooth, controllable strokes for the patterns, necessitating the robot to exert precise control over the dispensing. Furthermore,  most sauces exhibit high viscosity and exhibit time-dependent, nonlinear behavior, making precise dispensing control even more challenging. To tackle this challenge, we develop a nonlinear model to predict the liquid flow rate during a predefined squeezing motion. Additionally, we have constructed a model to forecast the relationship between flow rate and stroke width, enabling us to adjust the width of the drawn pattern based on the flow rate. These models are highly attuned to the properties of the liquid, particularly its viscosity, which we ascertain through a haptic exploration process.

\begin{figure}[t]
\centering
\includegraphics[width=0.95\linewidth]{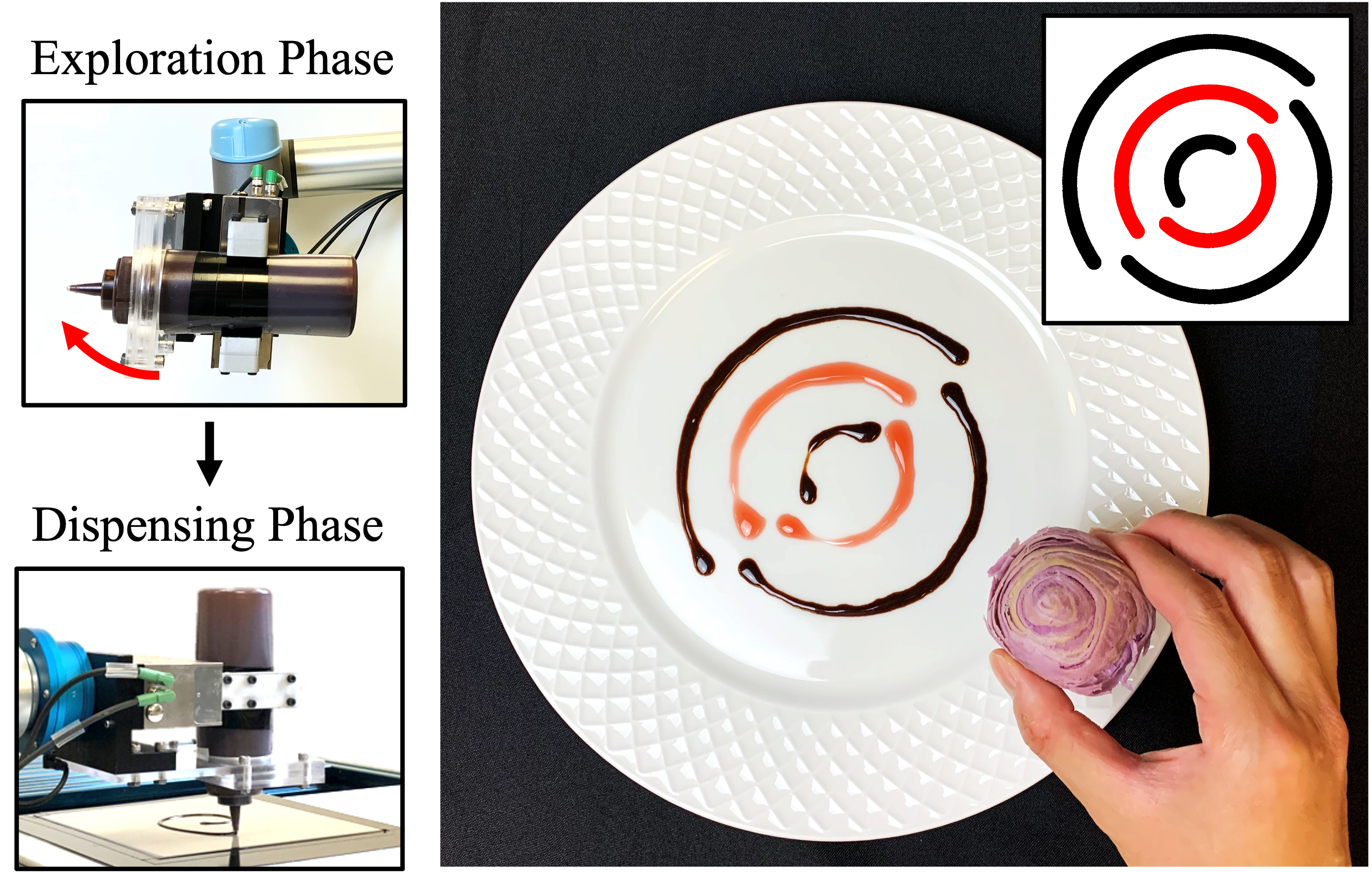}
\caption{We develop a liquid dispensing system capable of garnishing dishes with unseen sauces (chocolate and strawberry syrup here). The robot adjusts its dispensing strategy based on the sauce properties explored through a rotating action.}
\label{fig:fig1}
\end{figure}

In other words, we build a \textit{exploration $\rightarrow$ manipulation} pipeline for the liquid dispensing task. We use the haptic signal from a simple exploratory procedure as the input to predict the liquid flowing pattern during the squeezing motion with a neural network model and use the model output to control the robot’s moving speed for making a controlled-width stroke for drawing. Experiments show that with our method, a robot can make a stroke with an error of $28.8\%$ and variance of $8.4\%$ in width for unseen sauces and unknown fill levels. This enables the robot to use multiple sauces to robustly make an arbitrary drawing pattern on a plate. 

Our system will enhance the culinary capabilities of robots on both functional and artistic fronts. On the functional side, robots will be able to evenly distribute sauces or liquids across a pan or salad bowl. On the artistic side, our technology can craft intricate garnish patterns for dishes. Other than the potential of an immediate impact in culinary industry, our method can extend the robots’s involvement in various other industries, including pharmaceuticals, manufacturing, agriculture, and healthcare, where the precise liquid dispense control is crucial.

\section{Related Work}
\subsection{Liquid Manipulation}
Robot pouring is the most common task involving the manipulation of liquids \cite{Matl2019} \cite{do2018} \cite{kennedy2019} \cite{Lopez-Guevara2020}. Schenck and Fox \cite{Schenck2017} utilized RGB images to estimate liquid volume in containers for feedback control in pouring. 
Liang et al. \cite{Liang2019} used audio signals to predict liquid height in containers, which can be applied to various container types. Yamaguchi and Atkeson \cite{Yamaguchi2016} reconstructed liquid flow using optical flow methods and demonstrated its applicability to pouring tasks. While pouring has been extensively studied, viscous liquids are typically handled through squeezing actions. The most relevant literature about robot squeezing is robot pipetting \cite{Tegally2020} \cite{Kessey2022}, which focuses on dispensing precise small liquid volumes using micropipette technology, commonly employed in biological and chemical research. However, robot pipetting requires a custom pipetting system and is unsuitable for general-purpose kitchen tasks executed by robot arms. 

\subsection{Liquid Properties Estimation}
In contrast to pouring, squeezing is significantly influenced by liquid properties, such as viscosity or fill levels. To achieve effective squeezing, it is essential to perceive these liquid properties. Many works have studied liquid properties estimation. Huang et al. \cite{Huang2022} induced bottle perturbations through an impulse action. Based on the tactile data collected during the free oscillation period, they apply a physics-inspired approach to infer the liquid viscosities and fill levels precisely. Matl et al. \cite{Matl2019} collected haptics signals during bottle rotation and predicted the liquid properties using a physics-based approach. Saal et al. \cite{Saal10} actively selected optimal shaking frequencies and directions, utilizing perceived tactile signals to infer liquid viscosities. Similarly, we employ a rotational action and leverage the collected haptic signals to capture information about liquid properties. This information guides us in calculating the optimal drawing speed, ensuring precise control over the thickness and volume-per-length of the squeezed stream on a surface.

\subsection{Intuitive Physics}
Interacting with the environment to acquire physical parameters for control or prediction is known as intuitive physics \cite{Jiajun2015} \cite{Agrawal2017}. Wang et al. \cite{wang2021} tilted and shook objects to learn properties such as center of mass and moment of inertia, enabling precise control of swinging motions. Zeng et al. \cite{Zeng2020} focused on learning residual physics to execute tossing actions with arbitrary objects. The most relevant work to our work is \cite{Lopez-Guevara2020}. The authors stirred the liquid and optimized system parameters to match simulation results, facilitating tasks of pouring viscous liquids.
\section{Method} \label{method}

\begin{figure*}
\centering
\includegraphics[width=0.98\linewidth]{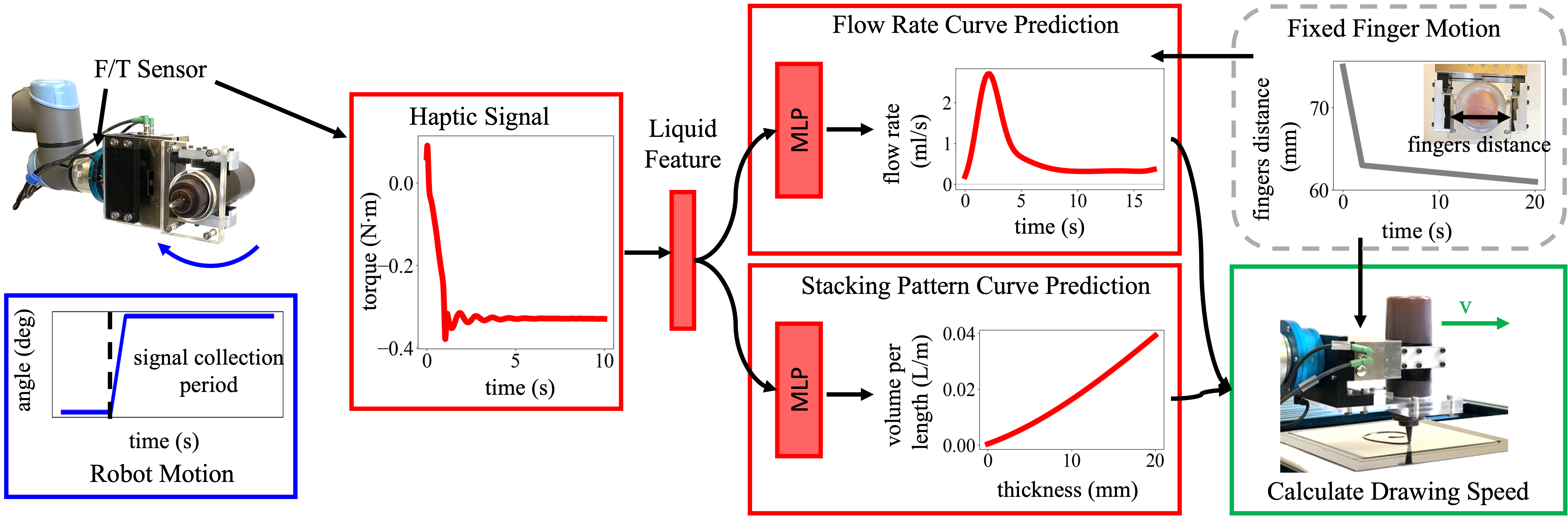}
\caption{Pipeline for controlling stream thickness and volume-per-length. The robot follows these steps to achieve its liquid dispensing goals: (1) Rotate the bottle $90$ degrees, pause for 10 seconds, and record haptic signals with a wrist-mounted F/T sensor to obtain the liquid feature. (2) The feature is fed into an MLP to predict the flow rate curve (flow rate over time) under the fixed finger motion and another MLP to predict the stacking pattern curve (volume-per-length for various stream thickness levels). (3) We utilize the predicted curves to guide the dispensing phase. The dispensing phase also uses the fixed finger motion. We calculate the required drawing speed to achieve the desired stream thickness and volume-per-length.}
\label{fig:fig2}
\end{figure*}

In this work, we aim to control the thickness and volume-per-length of the dispensed liquid stream from a squeezing bottle. To achieve this, we choose to keep the finger motion constant while adjusting the drawing speed (nozzle's movement speed). The required drawing speed varies depending on the type of liquid; for instance, viscous liquids require slower drawing speeds due to their lower flow rates. Therefore, we propose a two-phase approach: an exploration phase and a dispensing phase. In the exploration phase, we perturb the bottle to gather liquid property information. This information allows us to predict how the flow rate changes over time with our consistent squeezing motion and how the liquid forms strokes. In the dispensing phase, we utilize these predictions to guide the drawing speed, achieving control over stream thickness and volume-per-length. The pipeline is shown in Figure \ref{fig:fig2}.

During dispensing, our approach operates entirely in an open-loop manner. While it may seem intuitive to attach sensors for real-time measurement of liquid stream thickness or volume-per-length and feedback controlling the drawing speed, practical challenges arise. For example, visually monitoring the stroke is limited to scenarios where the dispensed liquid is neither occluded nor transparent and remains still. Similarly, monitoring the dispensed quantity using a force sensor on the robot's wrist is challenging due to sensor noise and motor vibrations, which can mask the subtle measured weight loss caused by the dispensed liquid.

\subsection{Haptic Exploration and Pre-processing}

In the exploration phase, the robot rotates the bottle to acquire information about the contained liquid's implicit properties that are relevant to flow rate and stroke width, such as viscosity and fill level. By rotating the bottle for $90$ degrees in one second, followed by a $10$-second pause, the robot's wrist-mounted F/T sensor records haptic signals (torque in the wrist's rotation direction) generated by the liquid's movement. According to \cite{Matl2019}, haptic signals during rotation provide information about fill level, while post-rotation torque changes indicate liquid viscosity. Figure \ref{fig:curves} (first column) shows haptic signals collected for three liquid types at varying fill levels.

We independently process the haptic signals recorded during and after the rotation, and subsequently concatenate them to create the liquid feature. We downsample the torque signals obtained during the one-second rotation period at intervals of $0.1$ seconds. For the torque signal collected during the subsequent free oscillation period, we extract its frequency domain within the range of $0.1$ Hz to $2.5$ Hz. These two features are concatenated with the total weight of the bottle and its liquid contents, resulting in a $33$ dimensional liquid feature.

\subsection{Inferring the Flow Rate Curve}

The \textit{flow rate curve} describes how the flow rate changes over time during squeezing. It is determined by both the liquid's property and the squeezing motion. In our work, we utilize a fixed gripper closing motion (Figure \ref{fig:fig2}): a $2$-second rapid squeeze followed by a $15$-second gradual squeeze. With this fixed squeezing motion, the flow rate curve can be entirely determined by the liquid properties, allowing it to be deduced solely from the liquid feature developed during the exploration phase. We train an MLP model taking the liquid feature as input to predict the flow rate curve. We avoid explicit predictions for liquid viscosity and filling level because other liquid properties encapsulated within the liquid feature may affect the flow rate curve. We parameterize the flow rate curve by the flow rate values at each second, and the MLP's objective is to predict these values accurately. The model is trained in a supervised fashion (see Section \ref{experiment_setup} for details). During testing, we connect the predicted flow rate values using splines to reconstruct the smooth flow rate curve. Figure \ref{fig:curves} shows the measured flow rate curve (second column) for three liquid types at varying fill levels, alongside the predicted flow rate curve (forth column) derived from the liquid feature.

\subsection{Inferring the Stacking Pattern Curve}

\begin{figure*}
\centering
\includegraphics[width=0.98\linewidth]{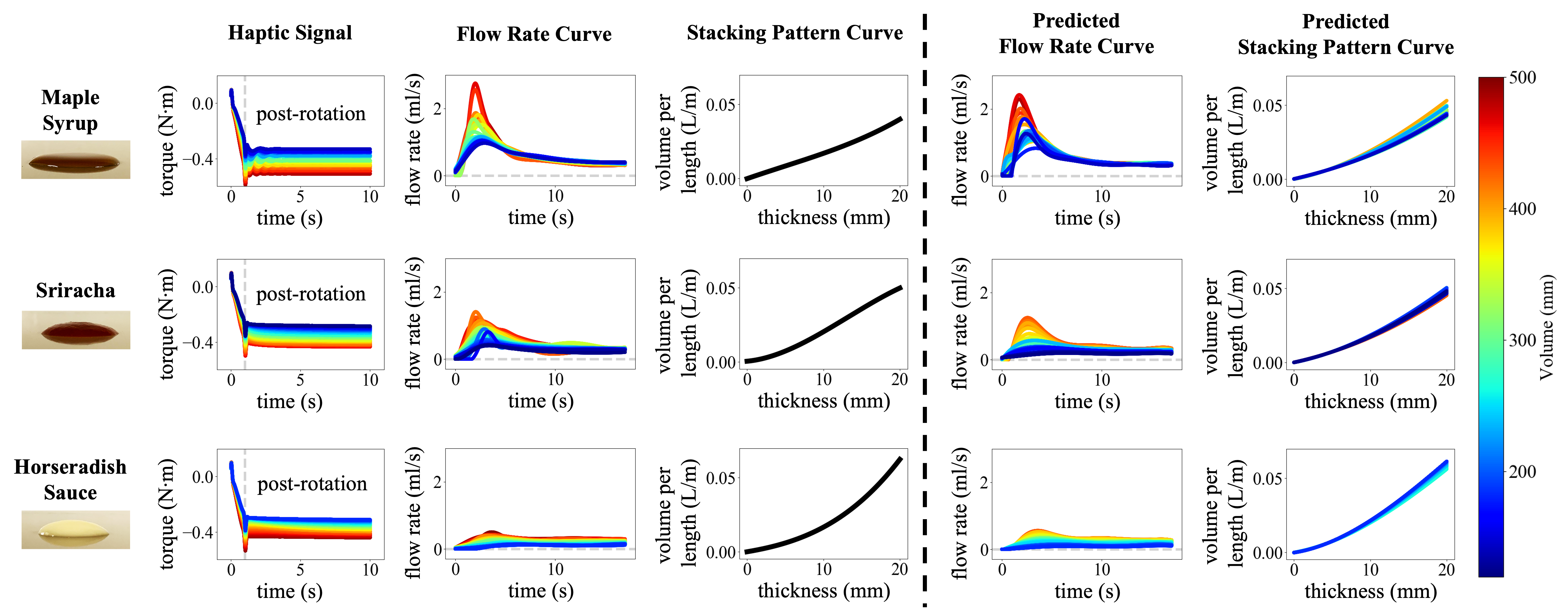}
\caption{(first three columns) The haptic signals collected in the exploration phase, the measured flow rate curves, and the measured stacking pattern curves of three testing liquids at different fill levels. These measured curves serve as ground truth. Maple syrup is the least viscous, while horseradish sauce is the most viscous. The side view of a $1$ ml drop of each liquid is shown. (last two columns) The predicted flow rate curves and the predicted stacking pattern curves based on the collected haptics signals.}
\label{fig:curves}
\end{figure*}

Additionally, we need to predict the liquid's \textit{stacking pattern curve} in order to control the geometric stream thickness. The stacking pattern curve describes how the volume-per-length value changes at different stream thicknesses. The curve is governed by the liquid viscosity. Viscous liquids tend to have a higher volume-per-length value for a fixed stream thickness, as it normally stacks higher. We train an MLP model taking the liquid feature as input to predict the stacking pattern curve of the liquid. To parameterize this curve, we use ten thickness values ranging from 5mm to 20mm, representing ten key points on the curve. The MLP will predict the volume-per-length of these ten points. During testing, we connect the predicted points with splines to reconstruct the smooth stacking pattern curve. Figure \ref{fig:curves} shows the measured stacking pattern curve (third column) for three liquid types, along with the predicted stacking pattern curve (fifth column) derived from the liquid feature.

\subsection{Control the Stream Thickness and Volume-per-length}

Based on the predicted flow rate curve and stacking pattern curve, we calculate the required drawing speed to achieve the desired stream thickness and volume-per-length. At any time during squeezing, we read the predicted liquid flow rate on the predicted flow rate curve. To reach a target stream with a volume-per-length of $\rho$ ml/cm while the current flow rate is predicted as $q$ ml/s, we simply adjust the drawing speed to $q/\rho$ cm/s. When our goal is to squeeze a stream with the desired thickness, we read the required volume-per-length for the stream on the predicted stacking pattern curve. Subsequently, we calculate the required drawing speed to achieve a stream with such volume-per-length.

\section{Experiments} \label{experiments}

In this section, we tested our method on the task of controlling the stream thickness with unseen liquid and unknown filling levels. We show that our system outperforms the baselines on this task. Since controlling volume-per-length is an intermediate goal of controlling stream thickness, our result also indicates our capability of controlling the stream's volume-per-length. 
We did not conduct specific experiments on volume-per-length control due to the difficulties of accurately measuring its ground truth. Finally, we demonstrate a robotics system to garnish dishes with unseen sauces.

\begin{figure}[t]
\centering
\includegraphics[width=0.95\linewidth]{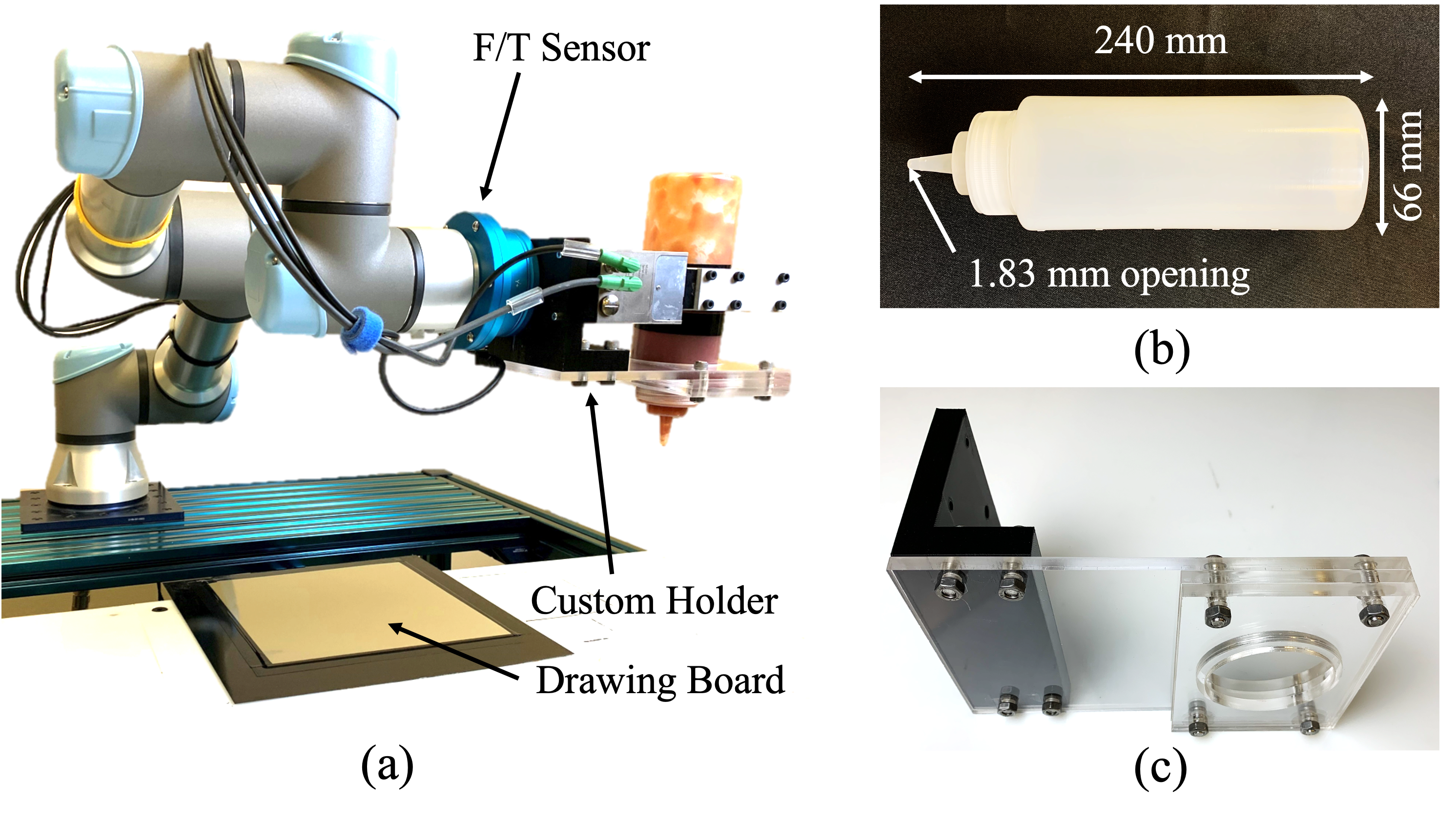}
\caption{(a) The F/T sensor is mounted at the robot wrist. The custom holder provides support to the bottle when the gripper is released. (b) The squeezing bottle. (c) The custom holder.}
\label{fig:setup}
\end{figure}

\begin{table}[ht]
\centering
\begin{tabular}{|c|c|c|c||c|}  
\hline 
\multicolumn{4}{|c||}{\textbf{Training}} & \textbf{Testing} \\ 
\hline 
\hline 
water & \shortstack{soy \\ sauce} & \shortstack{blueberry \\ syrup} & \shortstack{corn \\ oil} & \shortstack{maple \\ syrup}\\
\hline
\shortstack{heavy \\ cream} & \shortstack{strawberry \\ syrup} & \shortstack{butter \\ syrup} & kefir & \shortstack{chocolate \\ syrup}\\
\hline
\shortstack{orange \\ sauce} & \shortstack{katsu \\ sauce} & \shortstack{french \\ sauce} & \shortstack{fry \\ sauce} & sriracha \\
\hline
\shortstack{oyster \\ sauce} & ranch & ketchup & \shortstack{honey \\ bbq} & \shortstack{honey \\ mustard} \\
\hline
\shortstack{hoisin \\ sauce} & \shortstack{sour \\ cream} & \shortstack{char-shiu \\ sauce} & \shortstack{sweet-bean \\ sauce} & \shortstack{horseradish \\ sauce} \\
\hline
\end{tabular}
\caption{$20$ sauces in the training dataset (sorted from the least viscous to the most viscous) and $5$ sauces in the testing dataset (sorted from the least viscous to the most viscous).}
\label{tab:sauces}
\end{table}

\subsection{Experimental Setup} \label{experiment_setup}

\subsubsection{Robot Setup and Experimental Liquids}

The experiment setup is shown in Figure \ref{fig:setup}a. We use a 6DOF robot arm (UR5e by Universal Robotics) equipped with a 2-fingered gripper (WSG50 by Weiss Robotics) for our experiments. The fingers are made of aluminum. A 6-axis F/T sensor (NRS-6050-D80 by Nordbo Robotics) with $1000$ Hz sampling rate is attached to the robot wrist. To ensure the stability of the bottle while squeezing, a specially designed bottle holder (Figure \ref{fig:setup}c) is attached to the robot. All the liquids used in the experiments are contained within a $16$ oz squeezing bottle (Figure \ref{fig:setup}b). We select $25$ kinds of common liquids for squeezing (Table \ref{tab:sauces}), where $20$ is for training, and $5$ is for testing. The liquids cover a wide range of viscosity, ranging from $1$ cP (like water) to $70,000$ cP (like toothpaste). Note that high-viscosity liquids like honey can't be squeezed from our bottles. Before robot squeezing, we will manually shake the bottle to attain a homogeneous liquid state.

\subsubsection{Data Collection for Flow Rate Curve Prediction}
We collect data for flow rate curve prediction on each training liquid at various levels of bottle filling. Filling levels can span from $130$ ml ($25$\% full) to $500$ ml ($90$\% full). For each trial, we perform the rotation exploration action and the fixed squeezing motion. We collect the torque signal in the direction in which we rotate the wrist during the exploration phase. During the dispensing phase, we use a scale (USS-DBS28-50 by U.S. Solid) to measure the dispensed liquid weight in $5$ Hz. We then calculate the derivative of the measured weight over time and divide it by the true liquid density to obtain the flow rate curve. In total, $420$ data points are collected for training the MLP in predicting the flow rate curve.

\subsubsection{Data Collection for Stacking Pattern Curve Prediction}
The stacking pattern describes how the volume-per-length value changes at different stream thicknesses. This curve is specific to liquid type and doesn't depend on bottle filling levels. We excluded the six most watery liquids from the training set due to their inability to maintain their shape. For the remaining $14$ training liquids, we obtain their stacking pattern curve by squeezing twelve $10$ cm length streams on a flat surface, each with volume-per-length ranging from $0.08$ ml/cm to $0.5$ ml/cm. The thicknesses of each stream are manually labeled. Every data point represents a point on the stacking pattern curve for the liquid. The stacking pattern curve for each liquid is then obtained by connecting the twelve points using a spline.

\subsubsection{Neural Network Implementation}

We implement all MLPs with two hidden layers ($128$, $32$) in PyTorch and train them using Adam optimizer for $500$ epochs with a batch size of $32$. The initial learning rate is $0.005$, and it decreases by a factor of $0.9$ every $15$ epochs. Both the input and output of the MLP are normalized. The MLPs are trained in a supervised manner with an L1 loss function.

\subsection{Baseline Methods}
We implement three baseline methods to represent the squeezing performance under different robot settings and information levels:


\subsubsection{Simple (\textit{Simple})}
This approach intuitively assumes the volume reduction of the bottle during squeezing matches the dispensed liquid volume, with no consideration of liquid properties. It predicts the flow rate based on the volume reduction speed of the bottle. Instead of using our designed squeezing motion (quick squeeze followed by slow squeeze), we employ a consistent squeezing speed for this approach, which is more reasonable in this context. The stacking pattern curve is not available for this method.

\subsubsection{Physics Property Prediction (\textit{PP})}
This method utilizes the true liquid viscosity and true volume to predict the flow rate curve and the stacking pattern curve. In this case, the two MLPs take input as logarithmic viscosity and liquid volume. The liquid viscosities are measured by a viscometer (NDJ-5S by JIAWANSHUN). Note that this setting is under ideal assumption and not achievable for home robots. 

\subsubsection{Weight Feedback (\textit{WF})}
This approach directly monitors the dispensed volume using the F/T sensor attached to the robot's wrist. The reduction in gravity-directional force determines the dispensed weight, and the dispensed volume is calculated as the weight divided by true liquid density. We apply a Kalman Filter to reduce the influence of the sensor noise and motor vibrations.

\subsection{Flow Rate and Stacking Pattern Prediction}

\begin{table}[ht]
\centering
\begin{tabular}{|c|c|c|c|c|}  
\hline 
 & \textit{Ours} & \textit{Simple} & \textit{PP} & \textit{WF}\\ 
\hline 
\hline 
FC Error (ml/s)& 0.096 & 0.227 & \textbf{0.091} & 0.124 \\ 
\hline 
SC Error (ml/cm)& \textbf{0.039} & N/A & 0.045 & N/A \\ 
\hline 
\end{tabular}
\caption{Flow Rate Curve (FC) and Stacking Pattern Curve (SC) mean prediction error of testing liquids in various fill levels across different approaches.}
\label{tab:model_performance_result}
\end{table}

We compare our method to baseline models in predicting (estimating for \textit{WF}) the flow rate curve and stacking pattern curve for the testing liquids. The error metric calculates the average difference between predicted and true y-values at five evenly spaced points along both curves' x-axis. The result is shown in Table \ref{tab:model_performance_result}. Our method achieves a similar prediction error than \textit{PP}, which has additional access to the true liquid viscosity and fill level. Our approach outperforms the method that directly sensed the dispensed volume (\textit{WF}). The direct sensing method suffers from the sensor's noise (10 grams) compared to the average dispensed quantity of 5 grams, as well as interference from motor vibrations. The \textit{Simple} approach performs the worst, indicating that the reduction in bottle volume does not match the dispensed liquid volume. Figure \ref{fig:curves} shows the comparison between the curves predicted by our method and the actual curves for three of the test liquids.

\subsection{Result for Stroke Thickness Control}

\begin{figure}[t]
\centering
\includegraphics[width=0.95\linewidth]{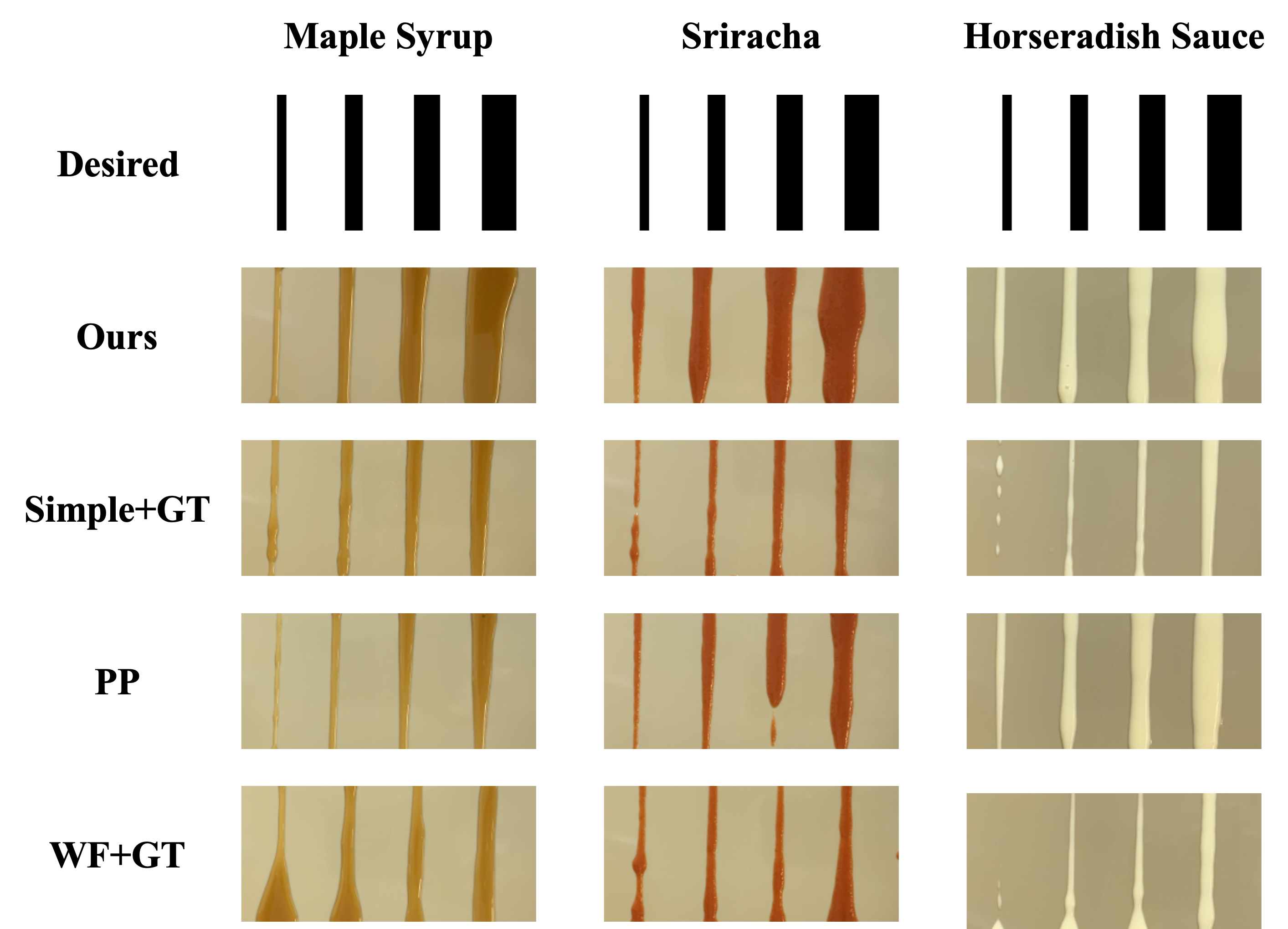}
\caption{Comparison between different approaches on drawing strokes with $5, 10, 15, 20$ mm thickness using three testing liquids at a volume of $400$ ml. The top row shows the desired stroke thickness.}
\label{fig:lines}
\end{figure}

\begin{table}[ht]
\centering
\begin{tabular}{|c|c|c|c|c|}  
\hline 
Thickness Metric & \textit{Ours} & \textit{Simple+GT} & \textit{PP} & \textit{WF+GT} \\ 
\hline 
\hline 
Error (mm)& \textbf{3.06} & 6.44 & 3.16 & 3.28\\ 
\hline 
Variance (mm)& 0.98 & \textbf{0.94} & 1.11 & 3.19 \\ 
\hline 
Percentage Error & 28.8\% & 50.2\% & \textbf{25.9\%} & 37.4\% \\ 
\hline 
Percentage Variance & \textbf{8.4\%} & 12.6\% & 13.3\% & 36.1\% \\ 
\hline 
\end{tabular}
\caption{Stream thickness error and variance of testing liquids across different methods.}
\label{tab:thickness_control}
\end{table}

\begin{figure}[!tbp]
\centering
\includegraphics[width=0.98\linewidth]{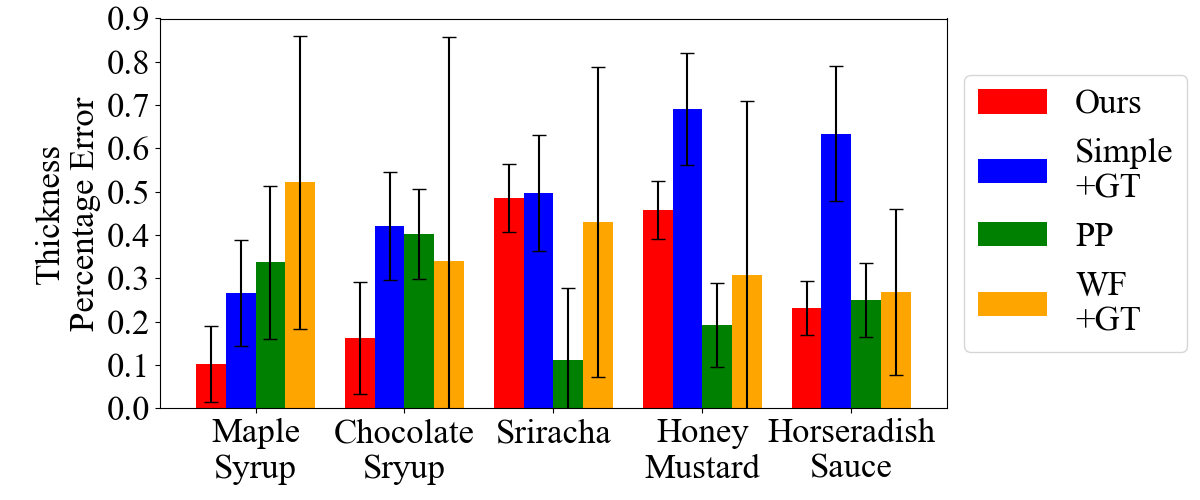}
\caption{Average stream thickness percentage error and percentage variance (shown as the error bar) on test liquids at various fill levels. The percentage variation indicates the consistency of stroke thickness throughout the line.}
\label{fig:thickness_control}
\end{figure}

In this experiment, we aim to control stream thickness when dispensed on a flat surface. We task the robot with squeezing streams of 10 cm in length at various filling levels of the testing liquids to achieve thicknesses of 5, 10, 15, and 20 mm. We assess accuracy by quantifying the deviation between the desired and actual thickness. Additionally, we calculate thickness consistency by measuring the standard deviation along the line; smaller deviations indicate more consistent thickness, which is preferable. We also compute percentage errors and percentage variances for each method. The results are shown in Figure \ref{fig:lines}, Table \ref{tab:thickness_control}, and Figure \ref{fig:thickness_control}. Since the baseline \textit{WF} and \textit{Simple} cannot predict stacking patterns, we incorporate the ground truth stacking pattern curve to control stream thickness. We denote this augmented baseline as \textit{WF+GT} and \textit{Simple+GT}.

In summary, our approach demonstrates comparable performance in both thickness accuracy and consistency control when compared to \textit{PP}, outperforming other baseline methods. While \textit{Simple+GT} can draw consistent strokes, it struggles to achieve the desired stroke width, particularly with viscous liquids. Conversely, \textit{WF+GT} achieves good accuracy in absolute thickness but encounters significant challenges in maintaining a consistent stroke due to motor vibrations affecting F/T sensor readings during arm movement.

\subsection{Dish Garnishing}

\begin{figure}[t]
\centering
\includegraphics[width=0.95\linewidth]{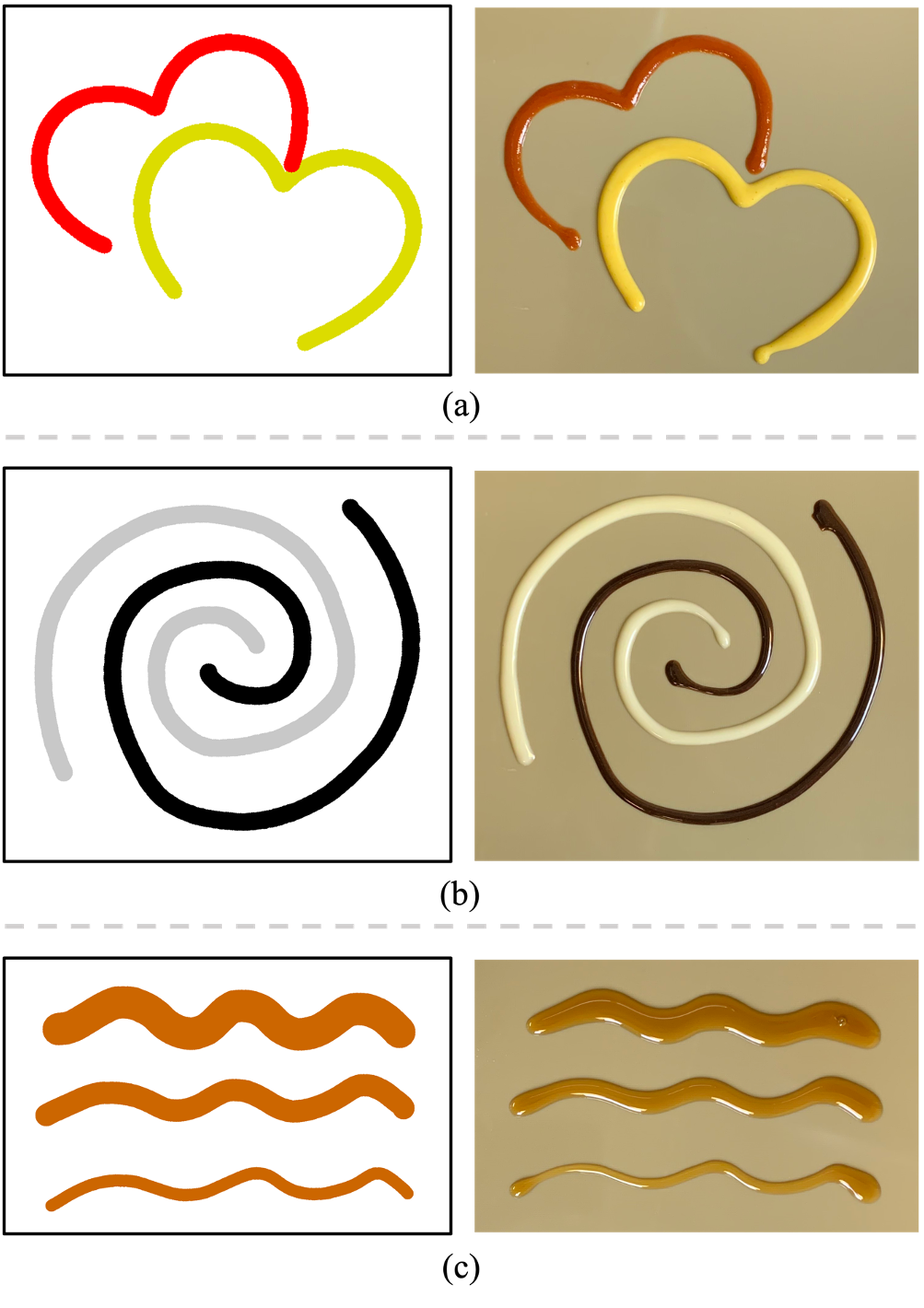}
\caption{Dish garnishing results: The left column shows the desired images and the right column shows the achieved drawings using our approach with the test liquids. (a) sriracha (red) + honey mustard (yellow). (b) horseradish sauce (white) + chocolate syrup (black). (c) maple syrup (brown).}
\label{fig:garnished}
\end{figure}

Finally, we task the robot with garnishing dishes by creating line art using testing liquids. We develop a robot motion trajectory directly from the provided drawing, and make the robot to proceed with the speed based on our model to maintain a desired width of the drawing stroke. Figure \ref{fig:garnished} shows the results. Despite the diverse liquid viscosities, our approach effectively controls stroke thickness with all testing sauces, achieving line art matching the desired drawings well.

\section{Discussion and Future Work} \label{future_work}
Today, the commercialization of autonomous kitchens has become a reality. Our autonomous dish garnishing system has substantial commercial potential as it can add artistic value to dishes. In the future, we plan to improve our system in a few ways. We aim to generalize our approach to accommodate sauce bottles with different nozzle sizes and volumes. While our approach is capable of creating line art using unseen sauces and can achieve good control over stroke thickness and consistency, it still exhibits a $29\%$ error on thickness control. We suspect this error stems from liquid properties such as surface tension and non-Newtonian behavior, which our current exploration methods cannot capture. We plan to develop new exploration methods to understand and handle these liquid properties. Finally, when visual monitoring of strokes is reliable in specific scenarios, we want to integrate it into our system to achieve even higher precision.
\section{Conclusion} \label{conclusion}
In this paper, we present a novel sauce dispensing method that offers precise control over stroke width and volume-per-length when squeezing on a surface. We also developed a dish garnishing system capable of handling various sauces. To adapt to sauces of different viscosities and fill levels, our method incorporates an exploration action to collect information about the liquid properties. With the collected information, we calculate the optimal drawing speed to achieve the desired stroke characteristics. In our experiments, we demonstrate the effectiveness of our approach in achieving precise stroke thickness control ($29\%$ error) and consistency ($8.4\%$ variance). It outperforms baseline methods, including one that directly monitors dispensed amounts using an F/T sensor. Moreover, we present qualitative results showing our method's superior performance when garnishing dishes with unseen sauces. To the best of our knowledge, we are the first to study robot squeezing actions using a general-purpose robot arm and gripper. We believe our dish garnishing system has broad applications, from assisting with cooking in households to potential commercial integration with existing autonomous kitchen systems.

\section*{ACKNOWLEDGMENT}
This work was supported by Toyota Research Institute.

\bibliographystyle{IEEEtran}
\bibliography{reference}

\end{document}